\documentclass[conference]{IEEEtran}
\IEEEoverridecommandlockouts
\usepackage{cite}
\usepackage{amsmath,amssymb,amsfonts}
\usepackage{algorithmic}
\usepackage{graphicx}
\usepackage{textcomp}
\usepackage{xcolor}
\usepackage{multirow}
\usepackage{arydshln}
\usepackage{todonotes}
\usepackage{etoolbox}
\usepackage{setspace}
\usepackage[bottom]{footmisc}
\usepackage{placeins}
\usepackage{wrapfig}
\usepackage[export]{adjustbox}

\def\BibTeX{{\rm B\kern-.05em{\sc i\kern-.025em b}\kern-.08em
    T\kern-.1667em\lower.7ex\hbox{E}\kern-.125emX}}
    
\usepackage{url}
\patchcmd{\section}{\centering}{}{}{}

\begin{document}
    
\title {On the Impact of Object and Sub-component Level Segmentation Strategies for Supervised Anomaly Detection within X-ray Security Imagery}

\author{\IEEEauthorblockN{Neelanjan Bhowmik$^1$, Yona Falinie A. Gaus$^1$, Samet Ak\c{c}ay$^1$, Jack W. Barker$^1$, Toby P. Breckon$^{1,2}$}
\IEEEauthorblockA{Department of \{Computer Science$^1$ $|$ Engineering$^2$\}, Durham University, UK}

}

\maketitle

\begin{abstract}
 X-ray security screening is in widespread use to maintain transportation security against a wide range of potential threat profiles.  Of particular interest is the recent focus on the use of automated screening approaches, including the potential anomaly detection as a methodology for concealment detection within complex electronic items. Here we address this problem considering varying segmentation strategies to enable the use of both object level and sub-component level anomaly detection via the use of secondary convolutional neural network (CNN) architectures. Relative performance is evaluated over an extensive dataset of exemplar cluttered X-ray imagery, with a focus on consumer electronics items. We find that sub-component level segmentation produces marginally superior performance in the secondary anomaly detection via classification stage, with true positive of $\sim98\%$ of anomalies, with a $\sim3\%$ false positive. 
\end{abstract}

    \section{Introduction} \label{sec:intro}
X-ray baggage security screening is widely used to maintain aviation and transport security, itself posing a significant image-based screening task for human operators reviewing compact, cluttered and highly varying baggage contents within limited time-scales. With both increased passenger throughput in the global travel network and an increasing focus on wider aspects of extended border security (e.g. freight, shipping, postal), this poses both a challenging and timely automated image classification task.

Prior work in the field has notably concentrated on the shaped-based detection of both threat and contraband (undeclared) items within X-ray imagery achieving both high detection performance with low false positive reporting \cite{Zheng2013:XrayVehicleThreat, Jaccard2014:XrayFreightContainer, Akcay2017:XrayICIP, Akcay2018:XrayIEEETransaction}. However, such approaches are insufficient when dealing with the detection of unknown anomalous items or materials potentially concealed within complex items such as consumer electronic devices. 
 
\begin{figure}[tb]
\centering
\includegraphics[width=\linewidth]{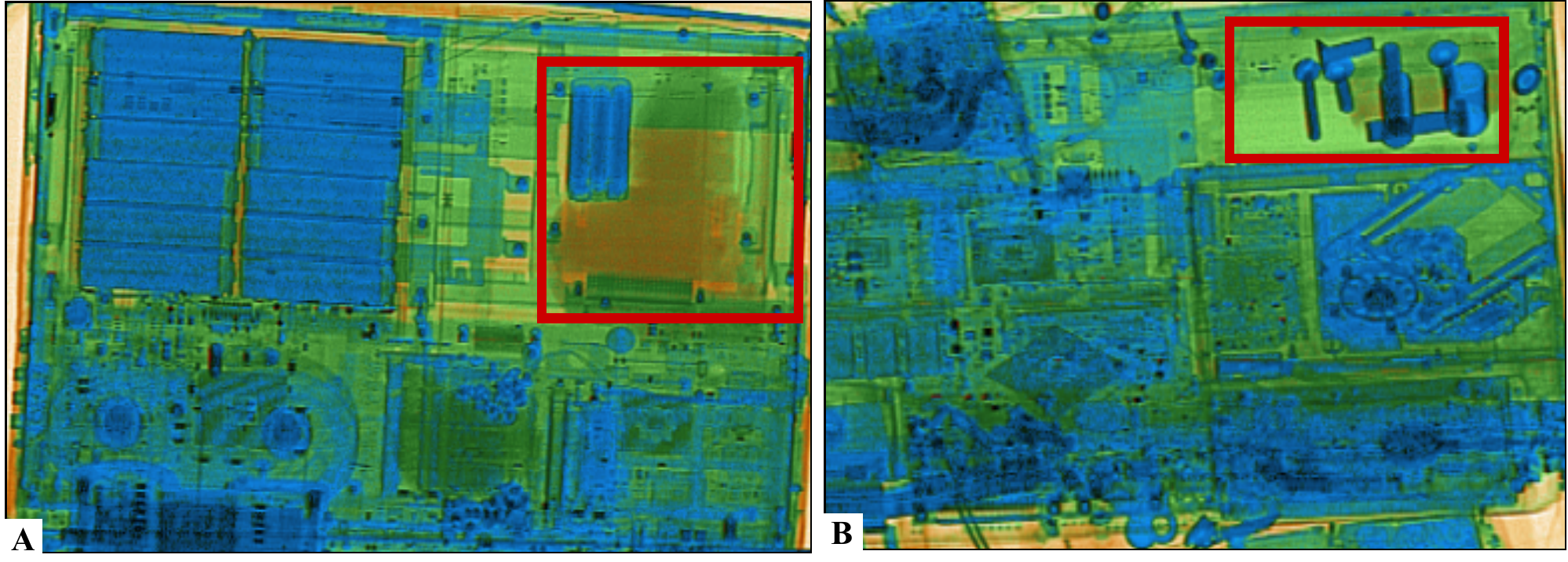}
\caption{Exemplar consumer electronics item within X-ray security imagery with both material (red box, A) and embedded object (red box, B) anomaly present.}
\label{fig:a_Exm_L}
\end{figure}

Whilst existing security scanners use dual-energy X-ray for materials discrimination, and highlight specific image regions matching existing threat material profiles \cite{singh2003explosives, wells2012review}, the detection of generalized anomalies within complex items remains challenging \cite{korupski2018evaluation} (e.g. Figure \ref{fig:a_Exm_L}). 

%Therefore, in this work, we side-step this issue by first using over segmentation method to a sub-component/structure level of the objects. Classification can then be performed looking at sub-component/structures and their relative positioning within the scanned item to identify anomalous sub-components within identified electronic/electrical items in cluttered X-ray imagery. For example, item such as laptop may contain anomalous sub-component which may unusual compared to other  

\begin{figure*}[!ht]
\centering
\includegraphics[width=\linewidth]{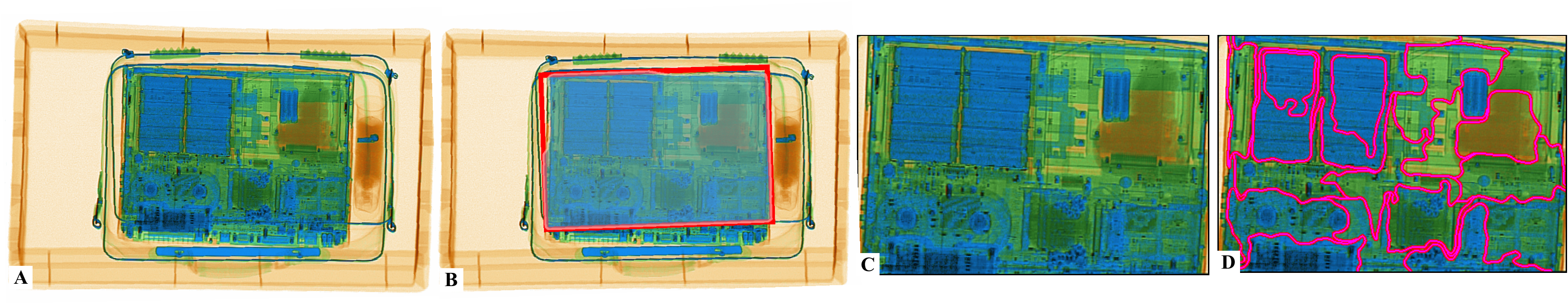}
\caption{Exemplar X-ray imagery (A) used for object level anomaly detection (B/C) via mask R-CNN segmentation and sub-component level anomaly detection (D) via superpixel over-segmentation.}
\label{fig:block_algo}
\end{figure*}

Within machine learning, anomaly detection involves learning a pattern or distribution of normality for a given data source and thus detecting significant deviations from this norm \cite{Patcha2007:AnomalyOverview}. Anomaly detection is an area of significant interest within computer vision, spanning biomedical imaging \cite{Schlegl2017:AnomalyGAN} to video surveillance \cite{Kiran2018:AnomalyVideo}. In our consideration of X-ray security imagery, we are looking for abnormalities that indicate concealment or subterfuge whilst working against a real-world adversary who may evolve their strategy to avoid detection. Such anomalies may present (or conceal) themselves within appearance space in the form of an unusual shape, texture or material density (i.e. dual energy X-ray colour) \cite{greenemeier_2010}. Alternatively they may present themselves in a semantic form, where the appearance of unfamiliar objects either globally or locally within the X-ray image \cite{brown_2018}.

Prior work on appearance and semantic anomaly detection, has considered unique feature representation as a critical component for detection within cluttered X-ray imagery \cite{Griffin2018:UnexpetedItem}. Early work on anomaly detection in X-ray security imagery \cite{Zheng2013:XrayVehicleThreat}, implements block-wise correlation analysis between two temporally aligned scanned X-ray images. More recently \cite{Andrews2016:AnomalyAutoEncoder}, anomalous X-ray items within freight containers have been detected using auto-encoder networks, and additionally via the use convolutional neural network (CNN) extracted features as a learned representation of normality across stream-of-commerce parcel X-ray images \cite{Griffin2018:UnexpetedItem}. In a similar vein, the work of \cite{Akcay2018:GANomaly} focuses on the use of a novel adversarial training architecture to detect anomalies as high reconstruction errors produced from a generator network adversarially trained on non-anomalous (benign) stream-of-commerce X-ray imagery only.

However, the majority of this prior anomaly detection work is focused at the image or object level, where anomaly presence is clear in appearance or semantic space, by asking the global question - \textit{is the image anomalous?}

These approaches \cite{Andrews2016:AnomalyAutoEncoder, Griffin2018:UnexpetedItem, Akcay2018:GANomaly}, fail to address the fact that anomaly presence maybe subtle and concealed (i.e. present) within a semantically benign object itself (e.g. Figure \ref{fig:a_Exm_L} A/B). In this case, we wish to ask a highly localised question - \textit{is this part of this complex object within the image anomalous ?}

In order to address this issue, we consider the task of image segmentation - if we first segment a class of object from the image, then potentially segment that object into its sub-components how well can this issue of subtle and concealed anomaly detection be addressed. 

To these ends, we introduce a side-by-side comparison of both object and sub-component level segmentation strategies for this case of intra-object anomaly detection. While anomaly detection at an object level is more common, the detection in sub-component level is still at infancy. The key concept is that whilst subtle localised anomalies maybe difficult to detect via an image level anomaly detection approach, we can instead target object level or sub-component level anomaly detection in isolation. Hence a more general learning-driven approach can developed at the object or sub-component level instead of tackling global signatures across  all possible objects - \textit{and thus being able to tell if they are anomalous or benign in appearance / semantic space}. 

Following the work in Zhang et al, \cite{Zhang2014:XrayCargo} where they leverage the use of superpixels \cite{Achanta2012:SLIC} within X-ray cargo image classification, we complement such approach with prior object segmentation \cite{He2017:MaskRCNN} as an enabler to sub-component level anomaly detection within X-ray security imagery. Using contemporary object segmentation via mask Region-based CNN (R-CNN) \cite{He2017:MaskRCNN} and Simple Linear Iterative Clustering (SLIC) \cite{Achanta2012:SLIC} superpixels, we evaluate alternate strategies for the detection of subtle intra-object anomalies at either a generalised object level or sub-component level segmentation strategy, thus facilitating effective anomaly detection independent of resolute object classification (Section \ref{sec:proposal}). Our work is evaluated over a range of large consumer electronics items with and without intra-object anomaly presence (Section \ref{sec:eva}). 

    \section{Proposed Approach}  \label{sec:proposal}
Our approach considers two automatic segmentation strategies for intra-object anomaly detection in X-ray security imagery (Sec. \ref{subs:segment}), as illustrated in the Figure \ref{fig:block_algo}A:- first, object level segmentation is performed (Figures \ref{fig:block_algo}B $\rightarrow{}$ \ref{fig:block_algo}C) and secondly, sub-component level segmentation is performed (Figure \ref{fig:block_algo}C $\rightarrow{}$ \ref{fig:block_algo}D). This is followed by secondary scale-specific variants to contemporary deep CNN architectures for final anomaly detection as a binary, $\{anomaly, benign\}$, classification task (Sec. \ref{subs:classify}).

\subsection{Segmentation Strategies}
\label{subs:segment}
% \textbf{Object Level Segmentation:} Following the mask R-CNN work of \cite{Ren2015:FasterRCNN}, our object level segmentation strategy performs classification and bounding box regression by sliding a small window of $n \times n$ network over the convolutional feature map output by the last shared  $1 \times 1$ convolution layer of a backbone CNN. In parallel, a mask branch utilizes a small fully connected network applied to each identified bounded object, predicting a segmentation mask in a pixel-to-pixel manner \cite{He2017:MaskRCNN}. Specifically, we use a FPN-50 backbone CNN, pre-trained from ImageNet and with all other parameterisation following \cite{He2017:MaskRCNN}. Our Mask R-CNN is applied to an input X-ray image (Fig. \ref{fig:block_algo}A with segmented object (Fig. \ref{fig:block_algo}B then isolated for subsequent anomaly classification (Fig. \ref{fig:block_algo}C).
\noindent \textbf{Object Level Segmentation:} Our first segmentation strategy builds upon the Faster R-CNN \cite{Ren2015:FasterRCNN} X-ray security image specific work of \cite{Akcay2018:XrayIEEETransaction}, to augment this model by adding two additional convolutional layers to construct a object boundary segmentation mask, following the Mask R-CNN concept of \cite{He2017:MaskRCNN}. This is performed by adding an additional branch to Faster R-CNN that outputs an additional image mask indicating pixel membership of a given detected object. Mask R-CNN \cite{He2017:MaskRCNN} also addresses feature map misalignment, found in Faster R-CNN \cite{Ren2015:FasterRCNN} for higher resolution feature map boundaries, via bilinear boundary interpolation.  Our Mask R-CNN is applied to an input X-ray image (Figure \ref{fig:block_algo}A) with segmented object (Figure \ref{fig:block_algo}B then isolated from the image for subsequent object level anomaly detection (Figure \ref{fig:block_algo}C).

\begin{figure}[tb]
\centering
\includegraphics[width=\linewidth]{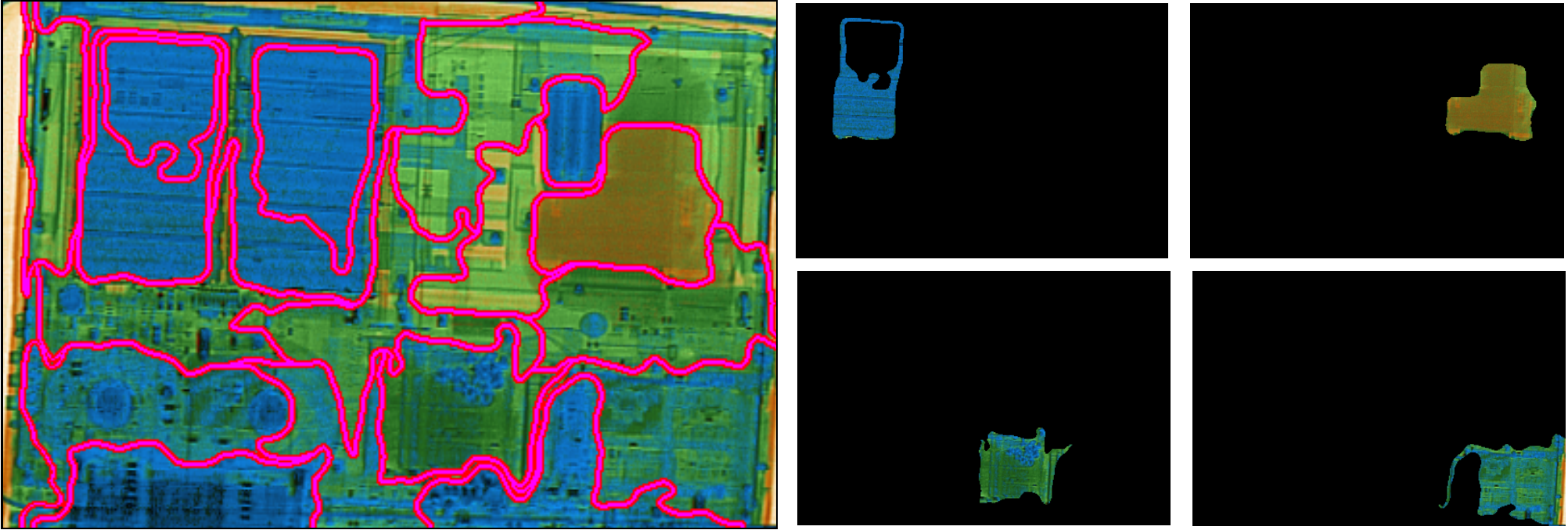}
\caption{Sub-component level object segmentation: each segment (pink contours) is extracted prior to CNN classification.}
\label{fig:seg}
\end{figure}

\noindent \textbf{Sub-component Level Segmentation:} Our second segmentation strategy uses image over-segmentation via Simple Linear Iterative Clustering (SLIC) \cite{Achanta2012:SLIC} superpixels. It performs iterative clustering in a similar manner to $k$-means, where the image is segmented into approximately equally-sized superpixels, whose total number $\textit{k}$ is user-defined. SLIC represents each pixel in $\mathbb{R}^5$, defined by the $\{L, a, b\}$ values of CIELAB colour space and the  $(x,y)$ pixel coordinate. Instead of using Euclidean distance, SLIC introduce a new distance measure that considers superpixel size. SLIC  takes as input a desired number of approximately equally-sized superpixel $K$, and for the images with $N$ pixels, with the approximate size of each superpixel will be $N$$/$$K$. Each of every approximately equally-sized superpixels, there will be a superpixel center at every grid interval  $S=\sqrt{N/K}$. Let $\big[l_i,a_i,b_i,x_i,y_i\big]^T$ be the five dimensional point of a pixel, cluster center $C_k$ should be in the same form as $\big[l_k,a_k,b_k,x_k,y_k\big]^T$. The distance measure $D_k$ is defined as:
\begin{equation}
\begin{split}
d_{lab}=\sqrt{(l_k-l_i)^2+(a_k-a_i)^2+(n_k-b_i)^2} \\
d_{xy}=\sqrt{(x_k-x_i)^2+(y_k-y_i)^2} \\
D_s=d_{lab}+\frac{m}{S}d_{xy}
\end{split}
\label{Eqn:superpixel}
\end{equation}
where $D_s$ is the sum of the $lab$ distance and the $xy$ plane distance $normalized$ by the grid interval $S$. Variable $m$ is introduced to control the compactness of the superpixel with the local convexity or concavity shape of each superpixel dependant on \textit{m} (low \textit{m} reduces the influence of coordinate information while for a high \textit{m} each superpixel will approximate a square shape). Our $m=20$ choice (by taking consideration of the size of the object present in an image), results in a set superpixel region conforming to convex and concave image shape boundaries as illustrated in the Figure \ref{fig:seg}. 

\subsection{Secondary Classification}
\label{subs:classify}
Each segmented image region, from object level or sub-component level segmentation, is subsequently classified using a deep CNN architecture model formulated as a binary, $\{anomaly, benign\}$, classification task. Three contemporary generalised CNN architectures plus leading fine-grain CNN classification approaches \cite{Lin2017BCNN, Zheng2017MA, Wang2018:DFLCNN}, specifically targeting the sub-categorization of pre-determined object types, are considered to form the basis of our anomaly detection study. 

\textit{VGG-16} \cite{Simonyan2014:VGGverydeep} is a seminal network architecture that consists of 16 deep convolutional layers, with a fixed kernel size of 3, stacked on top of each other in increasing depth.

\textit{SqueezeNet} \cite{Iandola2016:SqueezeNet} is a small network architecture that uses many 1-by-1 filters to aggressively reduce the number of weights. It offers equivalent accuracy to the AlexNet \cite{alexnet} yet operating with $50\times$ fewer parameters.

\textit{ResNet-50} \cite{He2015:ResNet} solves the issue of vanishing gradient present in the forward feed and backward propagation processing in previous CNN architectures by introducing skip connection, parallel to the regular convolutional layers numbering 50 in depth.

\textit{Fine-grain Classification} \cite{Lin2017BCNN, Zheng2017MA, Wang2018:DFLCNN} is put into effect as we can consider the task of anomaly detection in our case as a fine-grained image classification (FGIC) problem. The Xray screening imagery used, has very subtle differentiating factors in the sub-component of a given object (e.g laptop, bottle) and as such a fine grained approach should be used to detect finer class-specific discriminatory patches within objects \cite{Horn2015:Finegraindataset, Wegner2016:CatalogingPO}. \\
{\it Bilinear Convolutional Neural Network (BCNN)} \cite{Lin2017BCNN}  utilises a dual VGG-16 architecture in parallel with each stream implements uncommon, trivial elements of convolution and max pooling thus allowing focus on two separate distinct parts of the object. The two streams are concatenated into a bilinear vector using sum pooling over the outputs of both streams. This is then used in the final classification by feeding into the linear layers of the network, and finally a softmax layer to gain a probabilistic output of the most likely classifications for the image. \\
{\it Multi-Attention (MA)} \cite{Zheng2017MA} optimises part attentions of four distinct regions of an image using the feature channels in a VGG-16 architecture. This allows the network to focus on discriminative factors present in object parts, and use this in the final classification. Each of the four layers produces a classification at the end of the network linear layers which is then grouped by channel grouping loss in order to generate a final classification for a given object. \\
{\it Discriminative Filter Bank (DFL)} approach \cite{Wang2018:DFLCNN}, heightens the mid-level network in a VGG-16 based architecture, by learning a collection of $1 \times 1$ convolution filters known as a filter bank (FB), and a $92 \times 92$ with stride 8 to preserve global shape and appearance dependency in the image data \cite{Wang2018:DFLCNN}. These filters when properly initialised and successfully learned can respond to discriminative regions when convoluted over the image. 

When applied to the challenge of X-ray security screening, for the binary classification problem $\{anomaly, benign\}$, the models should be able to recognise much subtler visual differences and locations of such parts within object sub-components which will ultimately lead to more reliable classification.

% By using VGG16 \cite{Simonyan2014:VGGverydeep} the filters are added exactly at the $10^{th}$ convolutional layer, represents patches as small as $92 \times 92$ with stride 8, as stated in \cite{Wang2018:DFLCNN}, smatask.

%In this approach it enhance mid-level representation learning within the CNNs architecture, by learning a set of convolution filters, where such filters has been properly initialised and discriminatively learned in order to capture high quality discriminative patches. 
Each CNN architecture is trained via a transfer learning approach, with pre-training on the 1000-class ImageNet \cite{deng2009imagenet} object classification problem, for our final two-class (binary) X-ray imagery classification problem, $\{anomaly, \allowbreak benign\}$. For both object level and sub-component level segmentation our resulting image segments are padded and re-scaled to a common reference dimension (objects: $224 \times 224$; sub-components (superpixels): $190 \times 150$). Dataset imbalance, a common problem for anomaly detection problems where anomalous examples can be scarce and challenging to obtain, is addressed by up-sampling the anomalous class with the lesser volume of samples. In total training is performed over a dataset of 14,964 X-ray imagery ($70:30$ data split) and testing reported over a dataset of 7,878 X-ray imagery (50\%: anomalous and 50\%: benign) containing consumer electronics items.

Training is performed via transfer learning using stochastic gradient descent with a momentum of 0.9, a learning rate of 0.001, a batch size of 64 and categorical cross-entropy loss. All networks are trained on NVIDIA 1080 Ti GPU via PyTorch \cite{pytorch}.

 % \textcolor{red}{In addition to fine tuning, a layer freezing approach is adopted, by freezing the weights of the first few layers of the pre-trained network. The first few layers is essential because it capture universal features like curves and edges that are also relevant to our anomaly and benign classification problem.}
    \begin{table*}[ht]
\centering
\caption{Sub-component level segment classification of security X-ray imagery using varying CNN architectures.}
\begin{tabular}{llllllll}
\hline
Strategy & Model & Architecture &A & P & F1 & TP(\%) & FP(\%)\\ \hline \hline
\multirow{7}{*}{\shortstack[l]{Sub-component \\ level segmentation}}      & \multirow{4}{*}{\shortstack[l]{Binary Classification \\ via CNN}}  & ResNet-18 \cite{He2015:ResNet} & 97.10  &95.40   & 97.00 & 98.89 &4.69  \\      
&    & ResNet-50 \cite{He2015:ResNet} & 97.20   &95.50  & 97.10  &98.99 &4.54   \\ 
&    & SqueezeNet \cite{Iandola2016:SqueezeNet} &95.10   &92.60 &94.70 &\bf{99.10}  &8.90  \\ 
&   & VGG-16 \cite{Simonyan2014:VGGverydeep}  &93.70    &91.80  &93.30 &95.89  &8.55 \\ \cline{2-8}                   
 & \multirow{3}{*}{Fine-Grain Classification} 
 & BCNN \cite{Lin2017BCNN}       &97.54  &95.53  &97.49  &95.49  &4.30 \\
 & & MA \cite{Zheng2017MA}       &97.68  &95.81  &97.63  &96.32  &4.19 \\
 & & DFL \cite{Wang2018:DFLCNN}  &\bf{97.91}  &\bf{96.40}  &\bf{97.87}  & 98.20  &\bf{3.50} \\ \hline
 % & \multirow{1}{*}{Semi-supervised Classification}   & Ganomaly \cite{Akcay2018:GANomaly} & 49.64 & 49.71 & 49.09 & 60.11 & 60.82 \\ \hline

% \multirow{5}{*}{\shortstack[l]{Object level \\ segmentation}} & \multirow{4}{*}{\shortstack[l]{Binary Classification \\ via CNN}}  &ResNet-18 \cite{He2015:ResNet} & 86.20 & 80.60 & 76.90 & 95.42 & 21.13\\ 

%   &    & ResNet-50 \cite{He2015:ResNet} & 86.20   & \bf{84.50} & 84.50 & \bf{97.29} & 16.59  \\ 
%   &    & SqueezeNet \cite{Iandola2016:SqueezeNet} & 83.40  & 78.10  & 82.20 & 93.14 & 26.97 \\ 
%   &     & VGG-16 \cite{Simonyan2014:VGGverydeep} & 76.80  & 69.60 & 75.20 & 94.26 & 39.47 \\ \cline{2-8}
%   & \multirow{1}{*}{Fine-Grain Classification} & DFL \cite{Wang2018:DFLCNN} & \bf{89.77} & 83.70  & \bf{89.33} & 83.70 & \bf{03.88} \\  \hline

\end{tabular}
\label{Tab:class}
\end{table*}

\begin{table*}[ht]
\centering
\caption{Object level segment classification of security X-ray imagery using varying CNN architectures.}
\begin{tabular}{llllllll}
\hline
Strategy & Model & Architecture &A & P & F1 & TP(\%) & FP(\%)\\ \hline \hline
\multirow{5}{*}{\shortstack[l]{Object level \\ segmentation}} & \multirow{4}{*}{\shortstack[l]{Binary Classification \\ via CNN}}  &ResNet-18 \cite{He2015:ResNet} & 86.20 & 80.60 & 76.90 & 95.42 & 21.13\\ 

  &    & ResNet-50 \cite{He2015:ResNet} & 86.20   & \bf{84.50} & 84.50 & \bf{97.29} & 16.59  \\ 
  &    & SqueezeNet \cite{Iandola2016:SqueezeNet} & 83.40  & 78.10  & 82.20 & 93.14 & 26.97 \\ 
  &     & VGG-16 \cite{Simonyan2014:VGGverydeep} & 76.80  & 69.60 & 75.20 & 94.26 & 39.47 \\ \cline{2-8}
  & \multirow{1}{*}{Fine-Grain Classification} & DFL \cite{Wang2018:DFLCNN} & \bf{89.77} & 83.70  & \bf{89.33} & 83.70 & \bf{3.88} \\  \hline
% & \multirow{1}{*}{Semi-supervised Classification}  & Ganomaly \cite{Akcay2018:GANomaly}  & 60.61 60.00 @60.22 @68.88 @ 48.062 \\ \hline
\end{tabular}
\label{Tab:class_object}
\end{table*}

\section{Evaluation} \label{sec:eva}
%Our approach centre around the superpixel based localization approach which then feeds into variant CNN architecture. The approach is operating on X-ray security imagery, where it offer complete detection on anomaly and benign in the images.

Our evaluation considers the comparative performance of: (a) object level segmentation followed by anomaly detection via CNN classification (i.e., anomaly present in object as a whole - $\{anomaly, benign\}$) and (b) sub-component level segmentation followed by anomaly detection via CNN classification (i.e., anomaly present in image sub-component patches, i.e., superpixels - $\{anomaly, benign\}$). We consider statistical Accuracy \textit{(A)}, Precision \textit{(P)}, F-score \textit{(F1)}, True Positive \textit{(TP)} and False Positive \textit{(FP)} as presented in Tables \ref{Tab:class} and \ref{Tab:class_object}.

The X-ray security imagery dataset used for evaluation is obtained using a conventional 2D X-ray scanner with associated false colour materials mapping from dual-energy X-ray materials information \cite{mouton2015review}. It comprises large consumer electronics items (e.g., laptops) with and without intra-object anomaly concealment present. Anomaly concealments consist of marzipan, metal screws, metal plates, knife blades and similar inside the electronic items as illustrated in the examples of Figure \ref{fig:a_Exm_L} and Figure \ref{fig:block_algo}A/B. 

% removed - TPB, late 7/2/18 (any details should be in sec. 2.2)

%Classifier training was performed as set out in Sec. \ref{sec:proposal}. Each X-ray imagery is firstly run into object level segmentation in order to locate the electronic item inside the luggage/bag. The output will be candidate region of interest where it is cropped, flipped and scaled in order to enrich information given in X-ray imagery. These candidate regions of interest are trained using network set as discussed in the Sec. \ref{subs:classify}. For component level segmentation, the candidate region of interest is then over-segmented by approximately m=20  superpixels. 
% While the number superpixel is important, the shape of superpixel is also crucial because the size of the superpixel will determine the classification performance of CNN algorithm.
%  The shape of the superpixel may have convex and/or concave shape, as shown in Figure \ref{fig:seg}.
% In this part, the number of superpixel was determined empirically, but in future studies it may become possible to automate  this  number  by  performing a detailed analysis of the number of superpixels. Denoted as image patches, only the image patches is highlighted while the rest is zero, indicated as the background, as illustrated in Figure \ref{fig:seg}.

Performance evaluation of the object level segmentation and component level segmentation approaches are performed over a set of $7,878$ images annotated with ground truth anomaly location gathered using local access to a dual-view X-ray cabin baggage security scanner.

From the results presented in Tables \ref{Tab:class} and \ref{Tab:class_object}, we can observe that a sub-component level segmentation strategy, supported by the secondary fine-grain CNN classification of DFL model \cite{Wang2018:DFLCNN}, offers significantly superior anomaly detection performance (A: $97.91$, TP: $98.20$, FP: $3.50$ - Table \ref{Tab:class}) than an object level segmentation strategy overall (Table \ref{Tab:class_object}). Furthermore, fine-grain CNN classification similarly offers the highest overall accuracy and lowest false positive rate (A: $89.77$, FP: $3.88$ - Table \ref{Tab:class_object}) for object level segmentation. By contrast, the use of binary classification via a CNN offers superior performance for object level segmentation (Table \ref{Tab:class_object}) in terms of higher accuracy supported primarily by higher true positive detection at the expense of false positive reporting. Second stage binary classification via CNN performed less well overall with the sub-component segmentation strategy (lower accuracy (A) caused by significantly higher false positive (FP) - Table \ref{Tab:class_object}).

Fine grain classification model (DFL \cite{Wang2018:DFLCNN}) offer the lowest false positive and maximal accuracy for both segmentation strategies (Table \ref{Tab:class}). We can deduce that increased levels isolation via segmentation to the sub-component level improves the performance of the discriminative feature space learnt by the fine-grain technique \cite{Wang2018:DFLCNN,Zheng2017MA,Lin2017BCNN} whilst more classical object classification CNN architectures perform only marginally better on objects than sub-components (Table \ref{Tab:class_object}).

\begin{figure}[h]
\centering
\includegraphics[width=\linewidth]{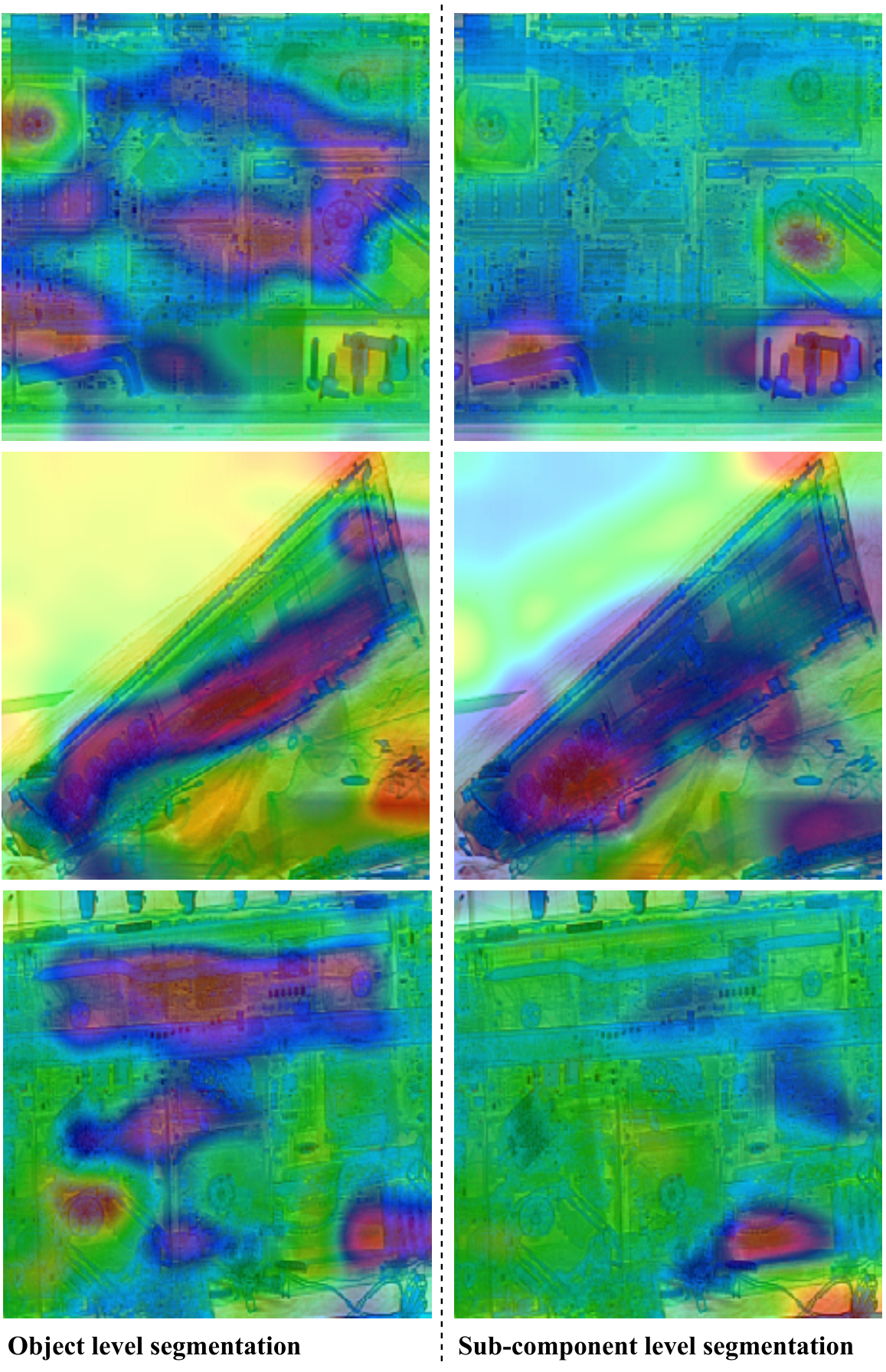}
\caption{Heatmap generated by convolutional activation map with DFL model \cite{Wang2018:DFLCNN} trained on object level segmentation images in first column and sub-component level segmentation images in second column. This shows where the model is looking (red/pink colour patch with primary focus regions) when selecting discriminative regions within the images using.}
\label{fig:cam_ex}
\end{figure}

Figure \ref{fig:cam_ex} illustrates the attention (red/pink patch with the highest focus) of the fine-grain DFL model \cite{Wang2018:DFLCNN} whilst trained on object-level and sub-component level segmentation data respectively. They are generated on the Rectified Linear Unit (ReLU) activations of the final layer before the fully connected layer in the VGG-16 \cite{Simonyan2014:VGGverydeep} architecture of DFL. It is evident when inspecting these that the sub-level components (Figure \ref{fig:cam_ex} second column) show attention over the anomalous parts of the laptops in each image while the object-level analysis (Figure \ref{fig:cam_ex} first column) shows relatively sporadic sparse attention over the images. This provides qualitative visualization supporting the performance of the sub-component level segmentation strategy outperforming object level segmentation. 

By enhancing the intermediate layer within the VGG-16 via a filter bank  \cite{Wang2018:DFLCNN}, we can hypothesise that this allows it to learn edge, corner and texture detail on specific sub-components at finer-level of the candidate region presented. As a result, the learned feature representation of anomaly against benign is highly discriminative leading to a significantly lower false positive than any other technique - at both the object and sub-component level (Tables \ref{Tab:class} and \ref{Tab:class_object}).

Binary classification via CNN using same VGG-16 architecture can achieve high true positive detection at the object level but at the expense of the highest false positive (TP: $94.25$, FP: $39.47$ - Table \ref{Tab:class_object}). This can also be observed for the ResNet and SqueezeNet architectures. For example, ResNet-50 achieves true positive of $97.29\%$, however suffering from high FP of $16.59\%$ for object level segment classification (Table \ref{Tab:class_object}). 
% We include the results with Ganomaly \cite{Akcay2018:GANomaly}, a semi-supervised approach, for both sub-component and object level segmentation in Tables \ref{Tab:class} and \ref{Tab:class_object}. However, semi-supervised approach fails to produce a satisfactory result due to the challenging nature of the dataset, and we believe it is not an equitable comparison between supervised and semi-supervised strategies.

Overall we observe that a sub-component level segmentation strategy, enabled via object segmentation via Mask R-CNN \cite{He2017:MaskRCNN} and subsequent superpixel over-segmentation via SLIC \cite{Achanta2012:SLIC}, consistently outperforms an object level segmentation strategy (via Mask R-CNN \cite{He2017:MaskRCNN} alone) when secondary region classification is performed using a specific fine-grain CNN variant \cite{Wang2018:DFLCNN}. The mean runtime for end-to-end $\{anomaly, benign\}$ classification strategy (object segmentation, followed by sub-component level segmentation and fine-grain classification) is $\sim500$ milliseconds, which is within the belt speed (0.2meter/second) of standard X-ray scanner \cite{gilardoni}.  

We primarily focus on supervised anomaly detection strategies and compared the performances amongst. Hence we do not include unsupervised or semi-supervised anomaly detection strategy \cite{Akcay2018:GANomaly} in our experiments and we believe it is not an equitable comparison between supervised and semi-supervised approaches.
To the best our knowledge, the proposed work on $\{anomaly, benign\}$ classification within large consumer electronics items, using sub-component level segmentation strategy, is first of its kind. As there is no prior related work is available on the literature of X-ray security imagery (e.g. sub-component level segmentation classification), we are unable to compare our strategies with any existing algorithm and present our results as the benchmark.

\begin{figure}[h]
\centering
\includegraphics[width=\linewidth]{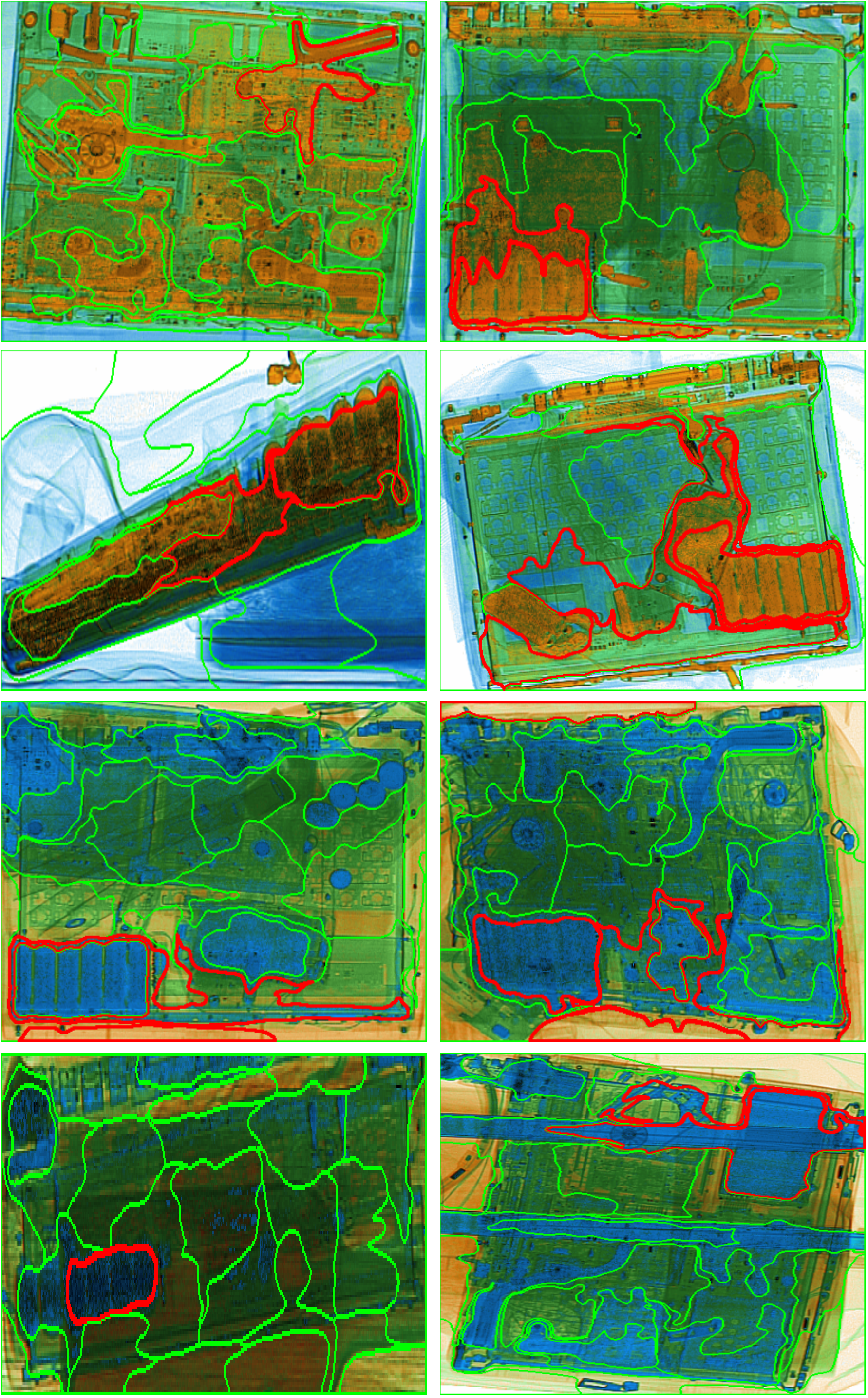}
\caption{Sub-component level segmentation via the use of SLIC approach \cite{Achanta2012:SLIC} and classification via fine-grain DFL \cite{Wang2018:DFLCNN} applied in X- ray security imagery (red contour: anomaly, green contour: benign).}
\label{fig:segEx}
\end{figure}

Figure \ref{fig:segEx} shows exemplar qualitative results of sub-component level segmentation with per superpixel classification using the fine-grained DFL \cite{Wang2018:DFLCNN} approach where we can see the colour coded set of anomalous (red) as well as benign (green) sub-component regions within the pre-isolated object-level image region. 

    \section{Conclusion} \label{sec:conclusion}
We assess the performance impact of varying segmentation strategies, such as object level and sub-component level segmentation, for intra-object anomaly detection within the context of X-ray security imagery. Our experimental comparison demonstrates the superiority of a sub-component level segmentation approach in combination with a specific fine-grain CNN architecture achieving a performance accuracy of $97.91\%$ with a notable $3.50\%$ false positive rate for realistic anomaly concealment within representative consumer electronic items. 

Future work will consider the conglomerate use of the multiple sub-component anomaly detection results in the robust determination of image-level anomaly vs. benign decision making for a broader range of object types.
%%%%%%%%%%%%%%%%%%%%%%%%%%%%%%%%%%%%%%%%%%%%%%%%%%%%%%%%%%%%%%%%
\\
\\
\noindent{\bf Acknowledgements}: Funding support - UK Department of Transport, Future Aviation Security Solutions (FASS) programme, (2018/2019).

    % \newpage

% \small{
\bibliographystyle{IEEEtran}
\bibliography{refs,icip-2016-refs}
% }
\end{document}